\documentclass[10pt,twocolumn,letterpaper]{article}

\usepackage{cvpr}
\usepackage{times}
\usepackage{epsfig}
\usepackage{graphicx}
\usepackage{amsmath}
\usepackage{amssymb}
\usepackage[pagebackref=true,breaklinks=true,letterpaper=true,colorlinks,bookmarks=false]{hyperref}
\usepackage{subfigure}
\usepackage{tabularx}
\usepackage{color}
\usepackage{multirow}
\usepackage{rotating}

\usepackage[breaklinks=true,bookmarks=false]{hyperref}

\cvprfinalcopy 

\def\cvprPaperID{****} 

\begin{document}

\title{Spatial Fusion GAN for Image Synthesis}

\author{Fangneng Zhan\\
Nanyang Technological University\\
50 Nanyang Avenue, Singapore 639798 \\
{\tt\small fnzhan@ntu.edu.sg}
\and
Hongyuan Zhu\\
Institute for Infocomm, A*STAR, Singapore\\
1 Fusionopolis Way, Singapore 138632\\
{\tt\small zhuh@i2r.a-star.edu.sg}
\and
Shijian Lu\\
Nanyang Technological University\\
50 Nanyang Avenue, Singapore 639798 \\
{\tt\small shijian.lu@ntu.edu.sg}
}

\maketitle

\begin{abstract}
Recent advances in generative adversarial networks (GANs) have shown great potentials in realistic image synthesis whereas most existing works address synthesis realism in either appearance space or geometry space but few in both. This paper presents an innovative Spatial Fusion GAN (SF-GAN) that combines a geometry synthesizer and an appearance synthesizer to achieve synthesis realism in both geometry and appearance spaces. The geometry synthesizer learns contextual geometries of background images and transforms and places foreground objects into the background images unanimously. The appearance synthesizer adjusts the color, brightness and styles of the foreground objects and embeds them into background images harmoniously, where a guided filter is introduced for detail preserving. The two synthesizers are inter-connected as mutual references which can be trained end-to-end without supervision. The SF-GAN has been evaluated in two tasks: (1) realistic scene text image synthesis for training better recognition models; (2) glass and hat wearing for realistic matching glasses and hats with real portraits. Qualitative and quantitative comparisons with the state-of-the-art demonstrate the superiority of the proposed SF-GAN.
\end{abstract}

\section{Introduction}
With the advances of deep neural networks (DNNs), image synthesis has been attracting increasing attention as a means of generating novel images and creating annotated images for training DNN models, where the latter has great potentials to replace the traditional manual annotation which is usually costly, time-consuming and unscalable. The fast development of generative adversarial networks (GANs) \cite{goodfellow2014} in recent years opens a new door of automated image synthesis as GANs are capable of generating realistic images by concurrently implementing a generator and discriminator. Three typical approaches have been explored for GAN-based image synthesis, namely, direct image generation \cite{mirza2014, radford2016, arjovsky2017}, image-to-image translation \cite{zhu2017, isola2017, liu2017, hoffman2018} and image composition \cite{lin2018, azadi2018}.

\begin{figure}[t]
\centering
\subfigure {\includegraphics[width=.19\linewidth,height=.15\linewidth]{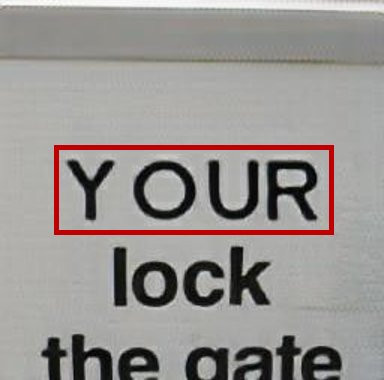}}
\subfigure {\includegraphics[width=.19\linewidth,height=.15\linewidth]{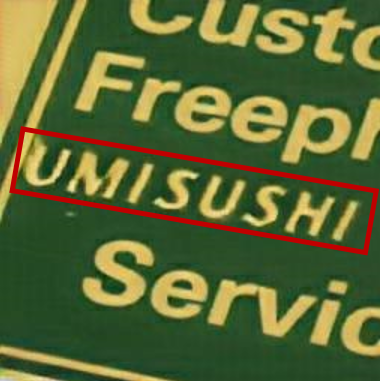}}
\subfigure {\includegraphics[width=.19\linewidth,height=.15\linewidth]{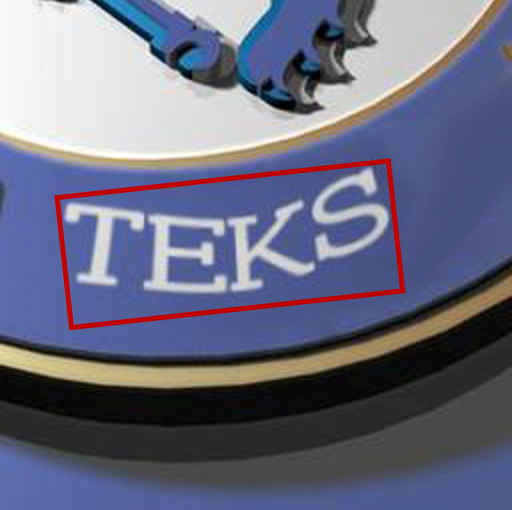}}
\subfigure {\includegraphics[width=.19\linewidth,height=.15\linewidth]{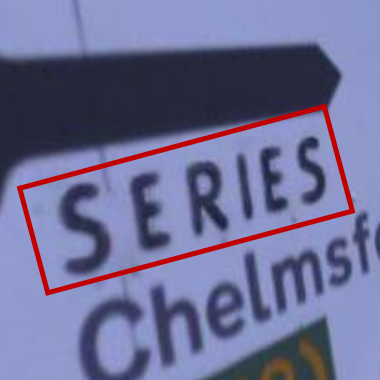}}
\subfigure {\includegraphics[width=.19\linewidth,height=.15\linewidth]{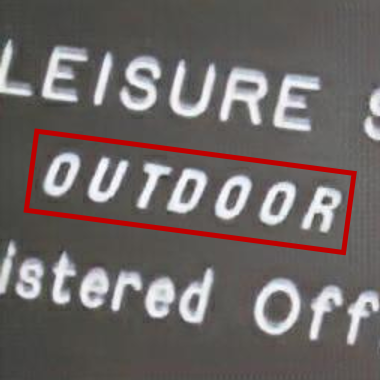}}
\vspace{-7.75 pt}

\subfigure {\includegraphics[width=.19\linewidth,height=.15\linewidth]{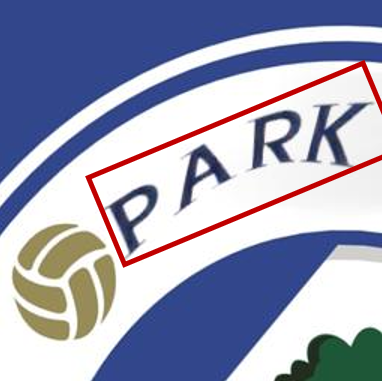}}
\subfigure {\includegraphics[width=.19\linewidth,height=.15\linewidth]{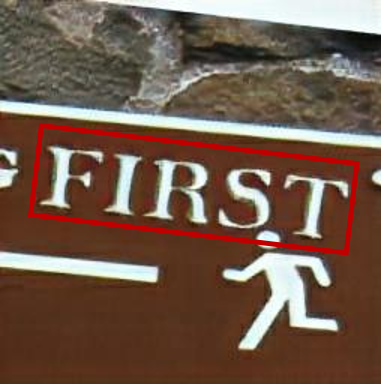}}
\subfigure {\includegraphics[width=.19\linewidth,height=.15\linewidth]{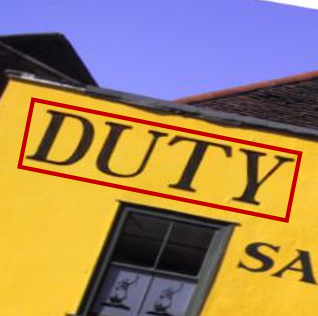}}
\subfigure {\includegraphics[width=.19\linewidth,height=.15\linewidth]{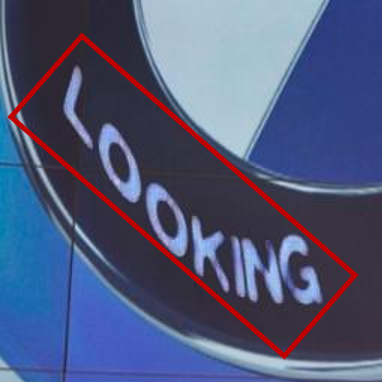}}
\subfigure {\includegraphics[width=.19\linewidth,height=.15\linewidth]{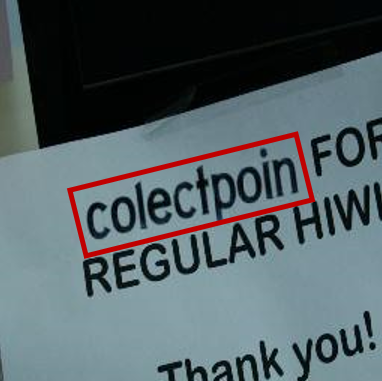}}
\vspace{-7.75 pt}

\subfigure {\includegraphics[width=.19\linewidth,height=.15\linewidth]{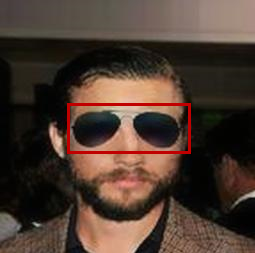}}
\subfigure {\includegraphics[width=.19\linewidth,height=.15\linewidth]{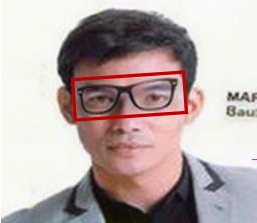}}
\subfigure {\includegraphics[width=.19\linewidth,height=.15\linewidth]{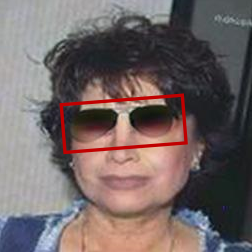}}
\subfigure {\includegraphics[width=.19\linewidth,height=.15\linewidth]{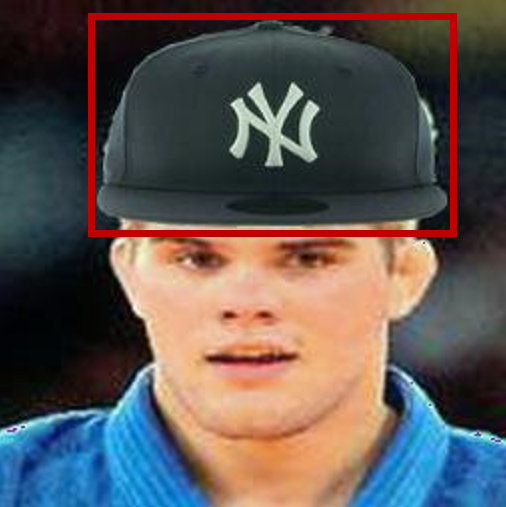}}
\subfigure {\includegraphics[width=.19\linewidth,height=.15\linewidth]{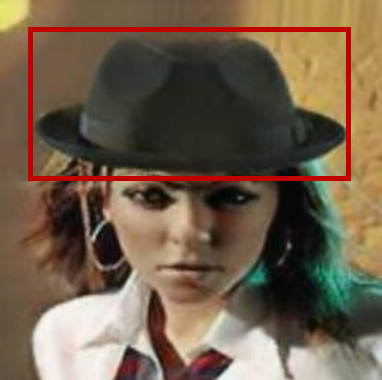}}
\vspace{0.01 pt}
\caption{The proposed SF-GAN is capable of synthesizing realistic images concurrently in geometric and appearance spaces. Rows 1 and 2 show a few synthesized scene text images and row 3 shows a few hat-wearing and glass-wearing images where the foreground texts, glasses and hats as highlighted by red-color boxes are composed with the background scene and face images harmoniously.}
\end{figure}

On the other hand, most existing GANs were designed to achieve synthesis realism either from geometry space or appearance space but few in both. Consequentially, most GAN-synthesized images have little contribution (many even harmful) when they are used in training deep network models. In particular, direct image generation still faces difficulties in generating high-resolution images due to the limited network capacity. 
GAN-based image composition is capable of producing high-resolution images \cite{lin2018, azadi2018} by placing foreground objects into background images. But most GAN-based image composition techniques focus on geometric realism only (e.g. object alignment with contextual background) which often produce various artifacts due to appearance conflicts between the foreground objects and the background images. As a comparison, GAN-based image-to-image translation aims for appearance realism by learning the style of images of the target domain whereas the geometric realism is largely ignored.

We propose an innovative Spatial Fusion GAN (SF-GAN) that achieves synthesis realism in both geometry and appearance spaces concurrently, a very challenging task in image synthesis due to a wide spectrum of conflicts between the foreground objects and background images with respect to relative scaling, spatial alignment, appearance style, etc. The SF-GAN address these challenges by designing a geometry synthesizer and an appearance synthesizer. The geometry synthesizer learns the local geometry of background images with which the foreground objects can be transformed and placed into the background images unanimously. A discriminator is employed to train a spatial transformation network, targeting to produce transformed images that can mislead the discriminator. The appearance synthesizer learns to adjust the color, brightness and styles of the foreground objects for proper matching with the background images with minimum conflicts. A guided filter is introduced to compensate the detail loss that happens in most appearance-transfer GANs. The geometry synthesizer and appearance synthesizer are inter-connected as mutual references which can be trained end-to-end with little supervision.

The contributions of this work are threefold. First, it designs an innovative SF-GAN, an end-to-end trainable network that concurrently achieves synthesis realism in both geometry and appearance spaces. To the best of our knowledge, this is the first GAN that can achieve synthesis realism in geometry and appearance spaces concurrently. Second, it designs a fusion network that introduces guided filters for detail preserving for appearance realism, whereas most image-to-image-translation GANs tend to lose details while performing appearance transfer. Third, it investigates and demonstrates the effectiveness of GAN-synthesized images in training deep recognition models, a very important issue that was largely neglected in most existing GANs (except a few GANs for domain adaptation \cite{ hoffman2018, isola2017, liu2017,zhu2017}). 

\section{Related Work}
\subsection{Image Synthesis}
Realistic image synthesis has been studied for years, from synthesis of single objects \cite{attias2016,park2015, su2015} to generation of full-scene images \cite{gaidon2016, richter2016}. Among different image synthesis approaches, image composition has been explored extensively which synthesizes new images by placing foreground objects into some existing background image. The target is to achieve composition realism by controlling the object size, orientation, and blending between foreground objects and background images. For example, \cite{gupta2016,jaderberg2014,zhan2018ver,zhan2019synth} investigate synthesis of scene text images for training better scene text detection \cite{xue2018acc} and recognition models \cite{zhan2019esir}. They achieve the synthesis realism by controlling a series of parameters such as text locations within the background image, geometric transformation of the foreground texts, blending between the foreground text and background image, etc. Other image composition systems have also been reported for DNN training \cite{dwibedi2017}, composition harmonization \cite{luan2018,tsai2017}, image inpainting \cite{zhao2018}, etc.

Optimal image blending is critical for good appearance consistency between the foreground object and background image as well as minimal visual artifacts within the synthesized images. One straightforward way is to apply dense image matching at pixel level so that only the corresponding pixels are copied and pasted, but this approach does not work well when the foreground object and background image have very different appearance. An alternative way is to make the transition as smooth as possible so that artifacts can be hidden/removed within the composed images, e.g. alpha blending \cite{uyttendaele2001}, but this approach tends to blur fine details in the foreground object and background images. In addition, gradient-based techniques such as Poisson blending \cite{perez2003} can edit the image gradient and adjust the inconsistency in color and illumination to achieve seamlessly blending.

Most existing image synthesis techniques aim for geometric realism by hand-crafted transformations that involve complicated parameters and are prone to various unnatural alignments. The appearance realism is handled by different blending techniques where features are manually selected and still susceptible to artifacts. Our proposed technique instead adopts a GAN structure that learn geometry and appearance features from real images with little supervision, minimizing various inconsistency and artifacts effectively.

\subsection{GAN}
GANs \cite{goodfellow2014} have achieved great success in generating realistic new images from either existing images or random noises. The main idea is to have a continuing adversarial learning between a generator and a discriminator, where the generator tries to generate more realistic images while the discriminator aims to distinguish the newly generated images from real images. Starting from generating MNIST handwritten digits, the quality of GAN-synthesized images has been improved greatly by the laplacian pyramid of adversarial networks \cite{denton2015}. This is followed by various efforts that employ a DNN architecture \cite{radford2016}, stacking a pair of generators \cite{zhang2017}, learning more interpretable latent representations \cite{chen2016}, adopting an alternative training method \cite{arjovsky2017}, etc.

Most existing GANs work towards synthesis realism in the appearance space. For example, CycleGAN \cite{zhu2017} uses cycle-consistent adversarial networks for realistic image-to-image translation, and so other relevant GANs \cite{isola2017, shrivastava2017}. LR-GAN \cite{jwyang2017} generates new images by applying additional spatial transformation networks (STNs) to factorize shape variations. GP-GAN \cite{wu2017} composes high-resolution images by using Poisson blending \cite{perez2003}. A few GANs have been reported in recent years for geometric realsim, e.g., \cite{lin2018} presents a spatial transformer GAN (ST-GAN) by embedding STNs in the generator for geometric realism, \cite{azadi2018} designs Compositional GAN that employs a self-consistent composition-decomposition network. 

Most existing GANs synthesize images in either geometry space (e.g. ST-GAN) or appearance space (e.g. Cycle-GAN) but few in both spaces. In addition, the GAN-synthesized images are usually not suitable for training deep network models due to the lack of annotation or synthesis realism. Our proposed SF-GAN can achieve both appearance and geometry realism by synthesizing images in appearance and geometry spaces concurrently. Its synthesized images can be directly used to train more powerful deep network models due to their high realism.

\subsection{Guided Filter}
Guided Filters \cite{he2015,he2013} use one image as guidance for filtering another image which has shown superior performance in detail-preserving filtering. The filtering output is a linear transform of the guidance image by considering its structures, where the guidance image can be the input image itself or another different image. 
Guided filtering has been used in various computer vision tasks, e.g., \cite{li2013} uses it for weighted averaging and image fusion, \cite{zhang2014} uses a rolling guidance for fully-controlled detail smoothing in an iterative manner, \cite{wu2017fast} uses a fast guided filter for efficient image super-solution, \cite{wliu2017} uses guided filters for high-quality depth map restoration, \cite{wliu2017_2} uses guided filtering for tolerance to heavy noises and structure inconsistency, and \cite{ham2018} puts Guided filtering as a nonconvex optimization problem and proposes solutions via majorize-minimization \cite{mm}.

Most GANs for image-to-image-translation can synthesize high-resolution images but the appearance transfer often suppresses image details such as edges and texture. How to keep the details of the original image while learning the appearance of the target remain an active research area. The proposed SF-GANs introduces guided filters into a cycle network which is capable of achieving appearance transfer and detail preserving concurrently.
\begin{figure*}
\centering
\includegraphics[width=0.985\linewidth,height=6cm]{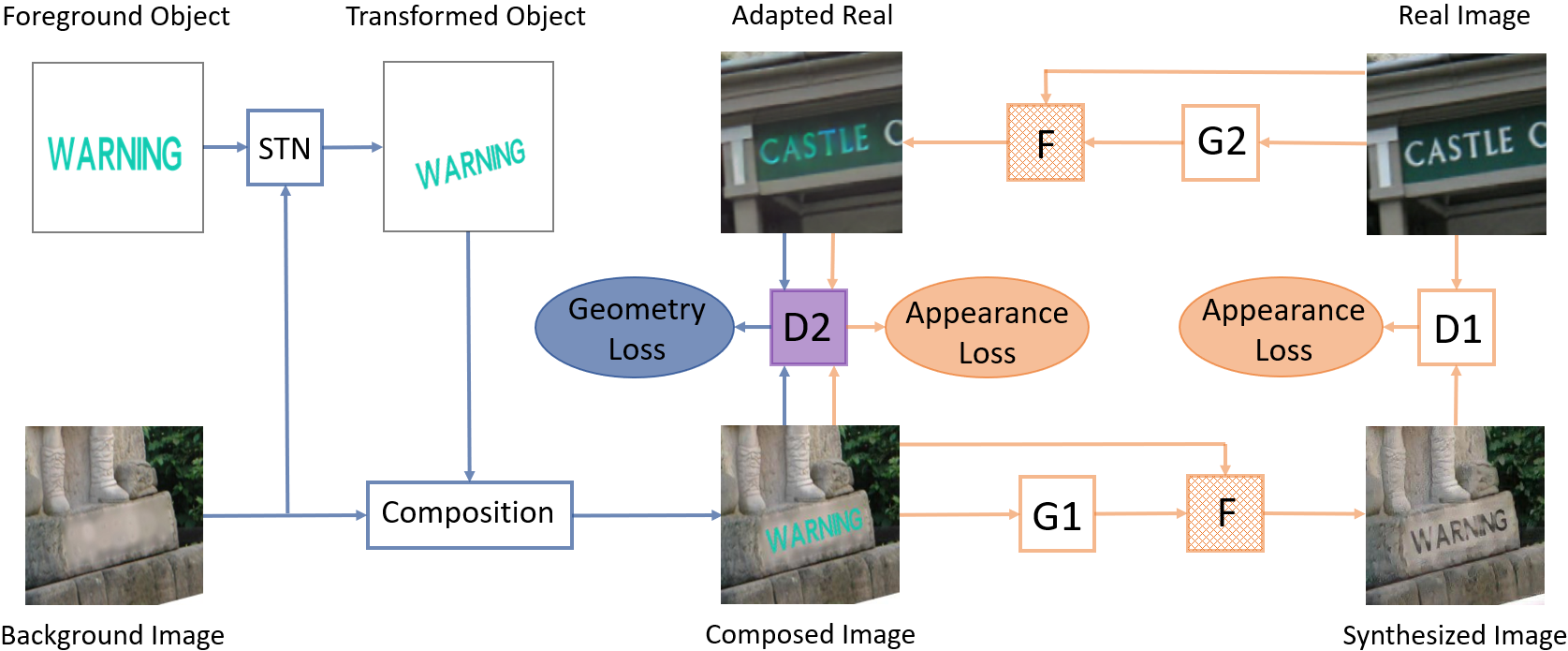}
\caption{The structure of the proposed SF-GAN: The geometry synthesizer is highlighted by blue-color lines and boxes on the left and the appearance synthesizer is highlighted by orange-color lines and boxes on the right. \textit{STN} denotes spatial transformation network, \textit{F} denotes guided filters, \textit{G1}, \textit{G2}, \textit{D1} and \textit{D2} denote the generators and discriminators. For clarity, cycle loss and identity loss are not included.}
\end{figure*}

\section{The Proposed Method}
The proposed SF-GAN consists of a geometry synthesizer and an appearance synthesizer, and the whole network is end-to-end trainable as illustrated in Fig. 2. Detailed network structure and training strategy will be introduced in the following subsections.

\subsection{Geometry Synthesizer}
The geometry synthesizer has a local GAN structure as highlighted by blue-color lines and boxes on the left of Fig. 2. It consists of a spatial transform network (STN), a composition module and a discriminator. The STN consists of an estimation network as shown in Table 1 and a transformation matrix which has $N$ parameters that control the geometric transformation of the foreground object. 

The foreground object and background image are concatenated to act as the input of the STN, where the estimation network will predict a transformation matrix to transform the foreground object. The transformation can be affine, homography, or thin plate spline \cite{tps} (We use thin plate spline for the scene text synthesis task and homography for the portrait wearing task).
Each pixel in the transformed image is computed by applying a sampling kernel centered at a particular location in the original image. With pixels in the original and transformed images denoted by $P^{s}=(p^{s}_{1},p^{s}_{2}, \dots, p^{s}_{N})$ and $P^{t}=(p^{t}_{1},p^{t}_{2}, \dots, p^{t}_{N})$, we use a transformation matrix \textit{H} to perform pixel-wise transformation as follows:
\begin{equation}
\left[
\begin{array}{l}
x^{t}_{i}
\\
y^{t}_{i}
\\
1
\end{array}
\right]
=
H
\left[
\begin{array}{l}
x^{s}_{i}
\\
y^{s}_{i}
\\
1
\end{array}
\right]
\end{equation}
\noindent where $p_{i}^{s} = (x_{i}^{s}, y_{i}^{s})$ and $p_{j}^{t} = (x_{i}^{t}, y_{i}^{t})$ denote the coordinates of the i-th pixel within the original and transformed image, respectively.
\begin{table}[t]
\caption{The structure of the geometry estimation network within the STN in Fig. 2}
\centering 
\begin{tabular}{|c|c|c|} 
\hline 
Layers & \multicolumn{1}{c|}{Out Size} & \multicolumn{1}{c|}{Configurations} \\
\hline
Block1 & $16 \times 50$ & $3 \times 3 \ conv, 32, 2 \times 2 \ pool$ \\
\hline
Block2 & $8 \times 25$ & $3 \times 3 \ conv, 64, 2 \times 2 \ pool$ \\
\hline
Block3 & $4 \times 13$ & $3 \times 3 \ conv, 128, 2 \times 2 \ pool$ \\
\hline
FC1 & $512$ & - \\
\hline
FC2 & N & - \\
\hline
\end{tabular}
\end{table}

The transformed foreground object can thus be placed into the background image to form an initially composed image (\textit{Composed Image} in Fig. 2). The discriminator \textit{D2} in Fig. 2 learns to distinguish whether the composed image is realistic with respect to a set of \textit{Real Images}. On the other hand, our study shows that real images are not good references for training geometry synthesizer.
The reason is real images are realistic in both geometry and appearance spaces while the geometry can only achieve realism in geometry space. The difference in appearance space between the synthesized images and real images will mislead the training of geometry synthesizer.
For optimal training of geometry synthesizer, the reference images should be realistic in the geometry space only and concurrently have similar appearance (e.g. colors and styles) with the initially composed images. Such reference images are difficult to create manually. In the SF-GAN, we elegantly use images from the appearance synthesizer (\textit{Adapted Real} shown in Fig. 2) as the reference to train the geometry synthesizer, more details about the appearance synthesizer to be discussed in the following subsection.

\subsection{Appearance Synthesizer}
The appearance synthesizer is designed in a cycle structure as highlighted in orange-color lines and boxes on the right of Fig. 2. It aims to fuse the foreground object and background image to achieve synthesis realism in the appearance space. Image-to-image translation GANs also strive for realistic appearance but they usually lose visual details while performing the appearance transfer. Within the proposed SF-GAN, guided filters are introduced which help to preserve visual details effectively while working towards synthesis realism within the appearance space.

\subsubsection{Cycle Structure}
The proposed SF-GAN adopts a cycle structure for mapping between two domains, namely, the composed image domain and the real image domain. Two generators \textit{G1} and \textit{G2} are designed to achieve image-to-image translation in two reverse directions, \textit{G1} from \textit{Composed Image} to \textit{Final Synthesis} and \textit{G2} from \textit{Real Images} to \textit{Adapted Real} as illustrated in Fig. 2. Two discriminator \textit{D1} and \textit{D2} are designed to discriminate real images and translated images.

In particular, \textit{D1} will strive to distinguish the adapted composed images (i.e. the \textit{Composed Image} after domain adaptation by \textit{G1}) and \textit{Real Images}, forcing \textit{G1} to learn to map from the \textit{Composed Image} to \textit{Final Synthesis} images that are realistic in the appearance space
G2 will learn to map from \textit{Real Images} to \textit{Adapted Real} images, the images that ideally are realistic in the geometry space only but have similar appearance as the \textit{Composed Image}. As discussed in the previous subsection, the \textit{Adapted Real} from \textit{G2} will be used as references for training the geometry synthesizer as it will better focus on synthesizing images with realistic geometry (as the interfering appearance difference 
has been compressed in the \textit{Adapted Real}).

Image appearance transfer usually comes with detail loss. We address this issue from two perspectives. The fist is by adaptive combination of cycle loss and identity loss. Specifically, we adopt a weighted combination strategy that assigns higher weight to the cycle-loss for interested image regions while higher weight to the identify-loss for non-interested regions. Take scene text image synthesis as an example. By assigning a larger cycle-loss weight and smaller identity-loss to text regions, it ensures a multi-mode mapping of the text style while keeping the background similar to the original image. The second is by introducing guided filters into the cycle structure for detail preserving, more details to be described in the next subsection.

\subsubsection{Guided Filter}
Guided filter was designed to perform edge-preserving image smoothing. It influences the filtering by using structures in a guidance image 
As appearance transfer in most image-to-image-translation GANs tends to lose image details, we introduce guided filters (\textit{F} as shown in Fig. 2) into the SF-GAN for detail preserving within the translated images. The target is to perform appearance transfer on the foreground object (within the \textit{Composed Image}) only while keeping the background image with minimum changes.
 \begin{figure}
\centering
\includegraphics[scale=0.2]{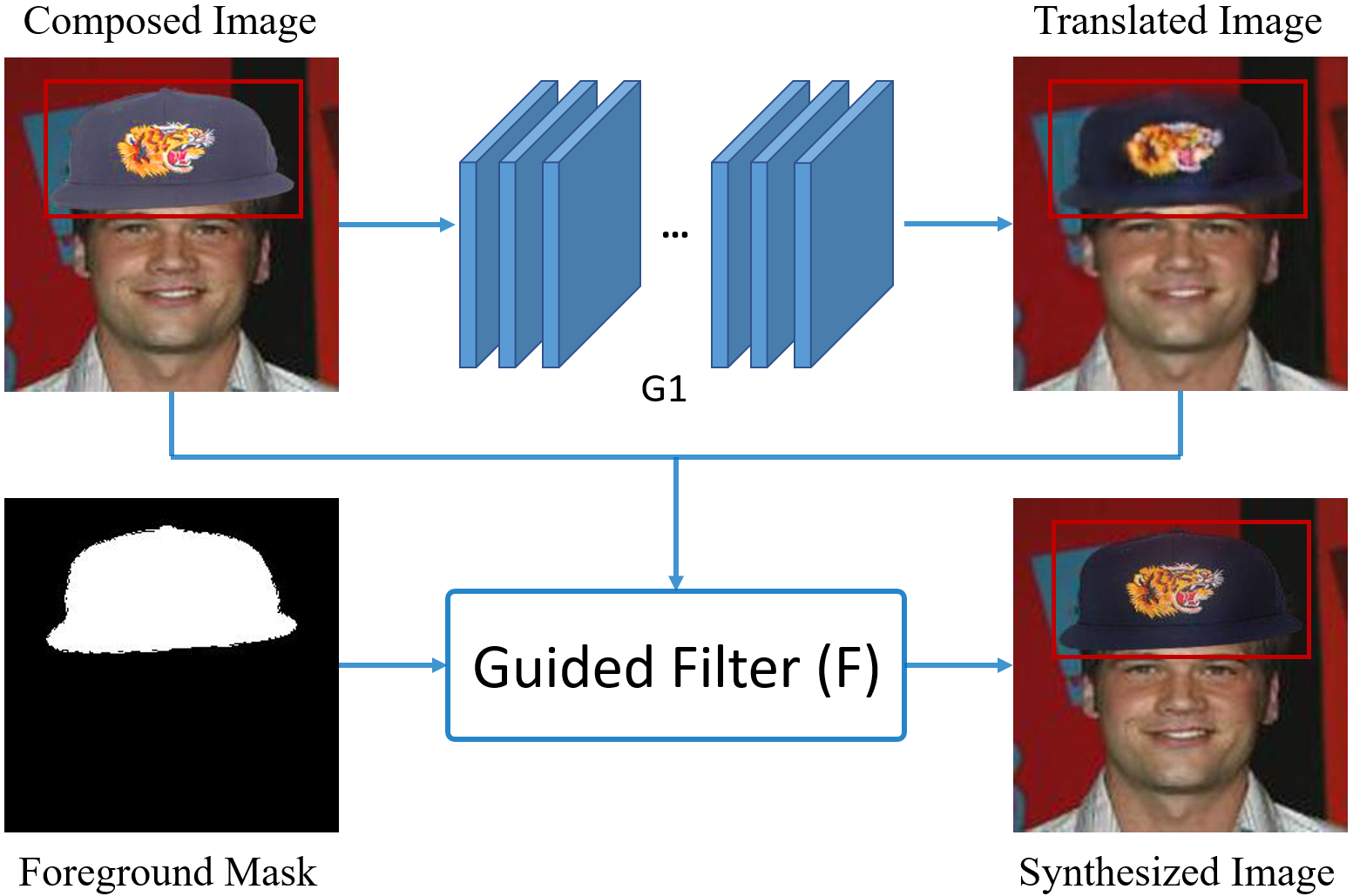}
\caption{Detailed structure of the guided filter \textit{F}: Given an image to be filtered (\textit{Composed Image} in Fig. 2), a translated image with smoothed details (the output of \textit{G1} in Fig. 2 where details are lost around background face and foreground hat areas) and the mask of foreground object hat (provided), \textit{F} produces a new image with full details (\textit{Synthesized Image}, the output of \textit{F} at the bottom in Fig. 2). It can be seen that the guided filter preserves details of both background image (e.g. the face area) and foreground hat (e.g. the image areas highlighted by the red-color box).}
\end{figure}

We introduce guided filters into the proposed SF-GAN and formulate the detail-preserving appearance transfer as a joint up-sampling problem as illustrated in Fig. 3. In particular, the translated images from the output of \textit{G1} (image details lost) is the input image $I$ to be filtered and the initially \textit{Composed Image} (image details unchanged) shown in Fig. 2 acts as the guidance image $R$ to provide edge and texture details. The detail-preserving image $T$ (corresponding to the \textit{Synthesized Image} in Fig. 2) can thus be derived by minimizing the reconstruction error between $I$ and $T$, subjects to a linear model:
\begin{equation}
T_{i} = a_{k}I_{i}+b_{k}, \forall i \in \omega_{k}
\end{equation}
where $i$ is the index of a pixel and $\omega_{k}$ is a local square window centered at pixel $k$. 

To determine the coefficients of the linear model $a_{k}$ and $b_{k}$, we seek a solution that minimizes the difference between $T$ and the filter input $R$ which can be derived by minimizing the following cost function in the local window:
\begin{equation}
E(a_{k},b_{k})=\sum_{i \in \omega_{k}}((a_{k}I_{i}+b_{k}-R_{i})^{2}+\epsilon a_{k}^2)
\end{equation}
where $\epsilon$ is a regularization parameter that prevents $a_{k}$ from being too large. It can be solved via linear regression:
\begin{equation}
a_{k}=\frac{
\frac{1}{\left| \omega \right|} 
\sum_{i \in \omega_{k}} I_{i}-\mu_{k}\overline{R}_{k}}
{\overline{\sigma}_{k}+\epsilon}
\end{equation}
\begin{equation}
b_{k}=\overline{R}_{k}-a_{k}\mu_{k}
\end{equation}
where $\mu_{k}$ and $\sigma_{k}^2$ are the mean and variance of $I$ in $\omega_{k}$, $\left| \omega \right|$ is the number of pixels in $\omega_{k}$, and $\overline{R}_{k}=\frac{1}{\left| \omega \right|} \sum_{i \in \omega_{k}}$ is the mean of $R$ in $\omega_{k}$.

By applying the linear model to all windows $\omega_{k}$ in the image and computing $(a_{k},b_{k})$, the filter output can be derived by averaging all possible values of $T_{i}$:
\begin{equation}
T_{i}=\frac{1}{\left| \omega \right|} 
\sum_{k:i \in \mu_{k}} (a_{k}I_{i}+b{k})
=\overline{a}_{i}I_{i}+\overline{b}_{i}
\end{equation}
where $\overline{a}_{i}=\frac{1}{\left| \omega \right|} \sum_{k \in \omega_{k}} a_{k}$ and $\overline{b}_{i}=\frac{1}{\left| \omega \right|} \sum_{k \in \omega_{i}} b_{k}$. We integrate the guide filter into the cycle structure network to implement an end-to-end trainable system.

\subsection{Adversarial Training}
The proposed SF-GAN is designed to achieve synthesis realism in both geometry and appearance spaces. The SF-GAN training therefore has two adversarial objectives, one is to learn the real geometry and the other is to learn the real appearance
The geometry synthesizer and appearance synthesizer are actually two local GANs that are inter-connected and need coordination during the training. For presentation clarity, we denote the \textit{Foreground Object} and \textit{Background Image} in Fig. 2 as the $x$, the \textit{Composed Image} as $y$ and the \textit{Real Image} as $z$ which belongs to domains $X$, $Y$ and $Z$, respectively.

For the geometry synthesizer, the \textit{STN} can actually be viewed as a generator $G0$ which predicts transportation parameters for $x$. After the transformation of the \textit{Foreground Object} and \textit{Composition}, the \textit{Composed Image} becomes the input of the discriminator $D2$ and the training reference $z^{'}$ comes from $G2(z)$ of the appearance synthesizer. For the geometry synthesizer, we adopt the Wasserstein GAN \cite{arjovsky2017} objective for training which can be denoted by:
\begin{equation}
\underset{G0}{min} \, \underset{D2}{max} \ E_{x \sim X} [D2(G0(x))] - E_{z^{'} \sim Z^{'}} [D2(z)]
\end{equation}
where $Z^{'}$ denotes the domains for $z^{'}$. 
Since \textit{G0} aims to minimize this objective against an adversary \textit{D2} that tries to maximize it, the loss functions of \textit{D2} and \textit{G0} can be defined by:
\begin{equation}
L_{D2} = E_{x \sim X }[D2(G0(x)] - E_{z^{'} \sim Z^{'}}[D2(z^{'})]
\end{equation}
\begin{equation}
L_{G0} = -E_{x \sim X}[D2(G0(x))]
\end{equation}

The appearance synthesizer adopts a cycle structure that consists of two mappings $G1: Y \rightarrow Z$ and $G2: Z \rightarrow Y$. It has two adversarial discriminators \textit{D1} and \textit{D2}.
D2 is shared between the geometry and appearance synthesizers, and it aims to distinguish $y$ from \textit{G2(z)} within the appearance synthesizer. The learning objectives thus consists of an adversarial loss for the mapping between domains and a cycle consistency loss for preventing the mode collapse. For the adversarial loss, the objective of the mapping $G1: Y \rightarrow Z$ (and the same for the reverse mapping $G2: Z \rightarrow Y$) can be defined by:
\begin{equation}
\begin{split}
L_{D1} = E_{y \sim Y }[D1(G1(y)] - E_{z \sim Z}[D2(z)]
\end{split}
\end{equation}
\begin{equation}
\begin{split}
L_{G1} = -E_{y \sim Y}[D1(G1(y))]
\end{split}
\end{equation}

As the adversarial losses cannot guarantee that the learned function maps an individual input $y$ to a desired output $z$, we introduce cycle-consistency, aiming to ensure that the image translation cycle will bring $x$ back to the original image, i.e. $y \rightarrow G1(y) \rightarrow G2(G1(y)) = y$. The cycle-consistency can be achieved by a cycle-consistency loss:
\begin{equation}
\begin{split}
L_{G1_{cyc}} = E_{y \sim p(y)}[\left \| G2(G1(y)) - y \right \|]
\end{split}
\end{equation}
\begin{equation}
\begin{split}
L_{G2_{cyc}} = E_{z \sim p(z)}[\left \| G1(G2(z)) - z \right \|]
\end{split}
\end{equation}
We also introduce the identity loss to ensure that the translated image preserves features of the original image:
\begin{equation}
\begin{split}
L_{G1_{idt}} = E_{y \sim Y} [\left \| G1(y) - y \right \|]
\end{split}
\end{equation}
\begin{equation}
\begin{split}
L_{G2_{idt}} = E_{z \sim Z} [\left \| G2(z) - z \right \|]
\end{split}
\end{equation}

For each training step, the model needs to update the geometry synthesizer and appearance synthesizer separately. In particular, $L_{D2}$ and $L_{G0}$ are optimized alternately while updating the geometry synthesizer. While updating the appearance synthesizer, all weights of the geometry synthesizer are freezed. In the mapping $G1: Y \rightarrow Z$, $L_{D1}$ and $L_{G1}+\lambda_{1}L_{G1_{cyc}}+\lambda_{2}L_{G1_{idt}}$ are optimized alternately where $\lambda_{1}$ and $\lambda_{2}$ controls the relative importance of the cycle-consistency loss and the identity loss, respectively.
In the mapping $G2: Z \rightarrow Y$, $L_{D2}$ and $L_{G2}+\lambda_{1}L_{G2_{cyc}}+\lambda_{2}L_{G2_{idt}}$ are optimized alternately.

It should be noted that the sequential updating is necessary for end-to-end training of the proposed SF-GAN. If discarding the geometry loss, we need update the geometry synthesizer according to the loss function of the appearance synthesizer. On the other hand, the appearance synthesizer will generate blurry foreground objects regardless of the geometry synthesizer and this is similar to GANs for direct image generation. As discussed before, the direct image generation cannot provide accurate annotation information and the directly generated images also have low quality and are not suitable for training deep network models.

\section{Experiments}
\subsection{Datasets}
\textbf{ICDAR2013} \cite{icdar2013} is used in the Robust Reading Competition in the International Conference on Document Analysis and Recognition (ICDAR) 2013. It contains 848 word images for network training and 1095 for testing.

\textbf{ICDAR2015} \cite{icdar2015} is used in the Robust Reading Competition under ICDAR 2015. It contains incidental scene text images that are captured without preparation before capturing. 2077 text image patches are cropped from this dataset, where a large amount of cropped scene texts suffer from perspective and curvature distortions.

\textbf{IIIT5K} \cite{iiit5k} has 2000 training images and 3000 test images that are cropped from scene texts and born-digital images. Each word in this dataset has a 50-word lexicon and a 1000-word lexicon, where each lexicon consists of a ground-truth word and a set of randomly picked words.

\textbf{SVT} \cite{wang2011} is collected from the Google Street View images that were used for scene text detection research. 647 words images are cropped from 249 street view images and most cropped texts are almost horizontal.

\textbf{SVTP} \cite{phan2013} has 639 word images that are cropped from the SVT images. Most images in this dataset suffer from perspective distortion which are purposely selected for evaluation of scene text recognition under perspective views.

\textbf{CUTE} \cite{risnumawan2014} has 288 word images mose of which are curved. All words are cropped from the CUTE dataset which contains 80 scene text images that are originally collected for scene text detection research.

\textbf{CelebA} \cite{liu2015} is a face image dataset that consists of more than 200k celebrity images with 40 attribute annotations. This dataset is characterized by large quantities, large face pose variations, complicated background clutters, rich annotations, and it is widely used for face attribute prediction.
\renewcommand\arraystretch{1.25}
\begin{table*}[t]
\caption{Scene text recognition accuracy over the datasets ICDAR2013, ICDAR2015, SVT, IIIT5K, SVTP and CUTE, where 1 million synthesized text images are used for all comparison methods as as listed.}
\hspace{0.5pt}
\centering 
\begin{tabular}{|l|p{1.8cm}<{\centering}|p{1.8cm}<{\centering}|p{1.2cm}<{\centering}|p{1.6cm}<{\centering}|p{1.4cm}<{\centering}|p{1.4cm}<{\centering}|p{1.6cm}<{\centering}|} 
\hline 
Methods & ICDAR2013 & ICDAR2015 & SVT & IIIT5K & SVTP & CUTE & AVERAGE \\
\hline
Jaderberg \cite{jaderberg2014} & 58.1 & 35.5 & 67.0 & 57.2 & \textbf{48.9} & 35.3 & 50.3 \\
Gupta \cite{gupta2016} & 62.2 & 38.2 & 48.8 & 59.1 & 38.9 & 36.3 & 47.3 \\
Zhan \cite{zhan2018ver} & \textbf{62.5} & 37.7 & 63.5 & 59.5 & 46.7 & 36.9 & 51.1 \\
ST-GAN \cite{lin2018} & 57.2 & 35.3 & 63.8 & 57.3 & 43.2 & 34.1 & 48.5  \\
\hline
\hline
SF-GAN(BS) & 55.9 & 34.9 & 64.0 & 55.4 & 42.8 & 33.7 & 47.8 \\
SF-GAN(GS) & 57.3 & 35.6 & 66.5 & 57.7 & 43.9 & 36.1 & 49.5 \\
SF-GAN(AS) & 58.1 & 36.4 & 66.7 & 58.5 & 45.3 & 35.7 & 50.1 \\
SF-GAN & 61.8 & \textbf{39.0} & \textbf{69.3} & \textbf{63.0} & 48.6 & \textbf{40.6} & \textbf{53.7} \\\hline
\end{tabular}
\end{table*}

\subsection{Scene Text Synthesis}
\noindent
\textbf{Data Preparation:} The SF-GAN needs a set of \textit{Real Images} to act at references as illustrated in Fig. 2. We create the \textit{Real Images} by cropping the text image patches from the training images of ICDAR2013~\cite{icdar2013}, ICDAR2015~\cite{icdar2015} and SVT~\cite{wang2011} by using the provided annotation boxes. While cropping the text image patches, we extend the annotation box (by an extra 1/4 of the width and height of the annotation boxes) to include certain local geometric structures

Besides the \textit{Real Images}, SF-GAN also needs a set of \textit{Background Images} as shown in Fig. 2. For scene text image synthesis, we collect the background images by smoothing out the text pixels of the cropped \textit{Real Images}. 
Further, the \textit{Foreground Object} (text for scene text synthesis) is computer-generated by using a 90k-lexicon. The created \textit{Background Images}, \textit{Foreground Texts} and \textit{Real Images} are fed to the network to train the SF-GAN.

For the training of scene text recognition model, texts need to be cropped out with tighter boxes (to exclude extra background). With the text maps as denoted by \textit{Transformed Object} in Fig. 2, scene text patches can be cropped out accurately by detecting a minimal external rectangle.
\newcommand{\tabincell}[2]{\begin{tabular}{@{}#1@{}}#2\end{tabular}}
\begin{figure}[t]
\begin{tabular}{p{1.2cm} p{1.2cm}<{\centering} p{1.2cm}<{\centering} p{1.2cm}<{\centering} p{1.2cm}<{\centering}}
\raisebox{-0.5\height}{\footnotesize{Foreground}} 
 & \raisebox{-0.5\height}{\includegraphics[width=1.25\linewidth,height=1.25\linewidth]{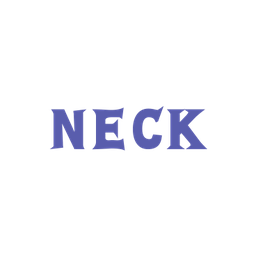}}
& \raisebox{-0.5\height}{\includegraphics[width=1.25\linewidth,height=1.25\linewidth]{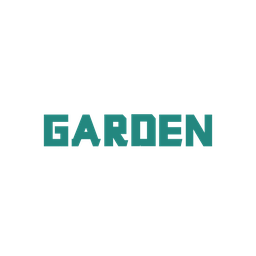}}
& \raisebox{-0.5\height}{\includegraphics[width=1.25\linewidth,height=1.25\linewidth]{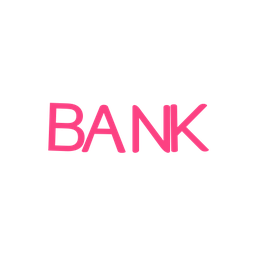}}
& \raisebox{-0.5\height}{\includegraphics[width=1.25\linewidth,height=1.25\linewidth]{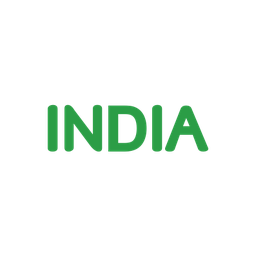}}
\\

\raisebox{-0.5\height}{\footnotesize{Background}} 
& \raisebox{-0.5\height}{\includegraphics[width=1.25\linewidth,height=1.25\linewidth]{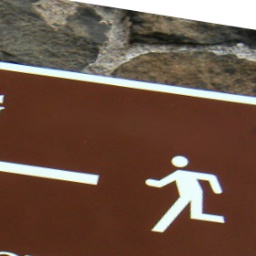}} 
& \raisebox{-0.5\height}{\includegraphics[width=1.25\linewidth,height=1.25\linewidth]{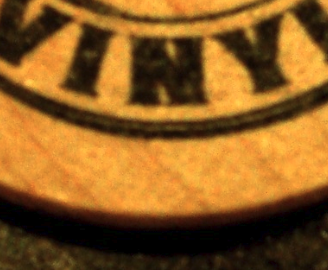}}
& \raisebox{-0.5\height}{\includegraphics[width=1.25\linewidth,height=1.25\linewidth]{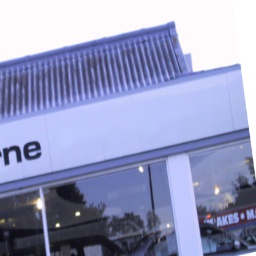}}
& \raisebox{-0.5\height}{\includegraphics[width=1.25\linewidth,height=1.25\linewidth]{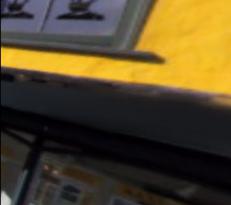}}
\vspace{-2.5 pt}
\\

\raisebox{-0.5\height}{\footnotesize{ST-GAN}} 
& \raisebox{-0.5\height}{\includegraphics[width=1.25\linewidth,height=1.25\linewidth]{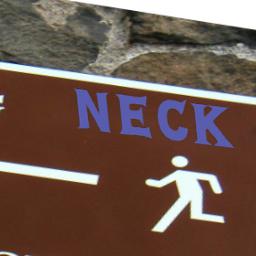}} 
& \raisebox{-0.5\height}{\includegraphics[width=1.25\linewidth,height=1.25\linewidth]{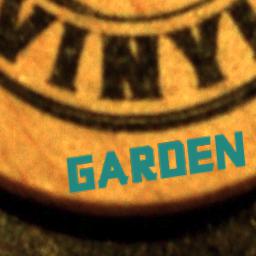}}
& \raisebox{-0.5\height}{\includegraphics[width=1.25\linewidth,height=1.25\linewidth]{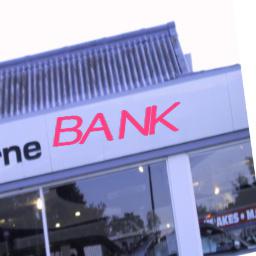}}
& \raisebox{-0.5\height}{\includegraphics[width=1.25\linewidth,height=1.25\linewidth]{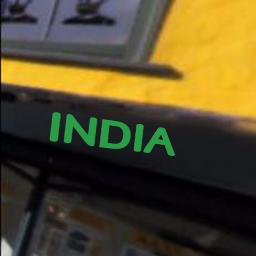}}
\vspace{-2.5 pt}
\\

\raisebox{-0.5\height}{\footnotesize{CycleGAN}} 
& \raisebox{-0.5\height}{\includegraphics[width=1.25\linewidth,height=1.25\linewidth]{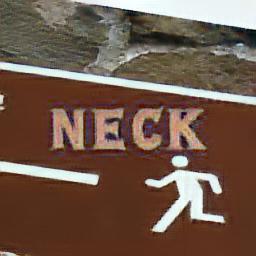}} 
& \raisebox{-0.5\height}{\includegraphics[width=1.25\linewidth,height=1.25\linewidth]{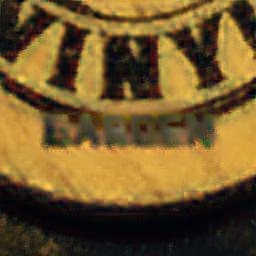}}
& \raisebox{-0.5\height}{\includegraphics[width=1.25\linewidth,height=1.25\linewidth]{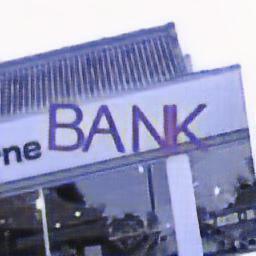}}
& \raisebox{-0.5\height}{\includegraphics[width=1.25\linewidth,height=1.25\linewidth]{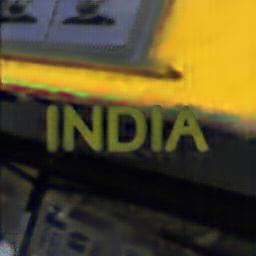}}
\vspace{-2.5 pt}
\\

\raisebox{-0.5\height}{\footnotesize{SF-GAN(GS)}}
& \raisebox{-0.5\height}{\includegraphics[width=1.25\linewidth,height=1.25\linewidth]{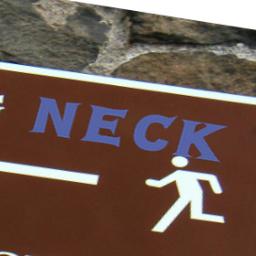}} 
& \raisebox{-0.5\height}{\includegraphics[width=1.25\linewidth,height=1.25\linewidth]{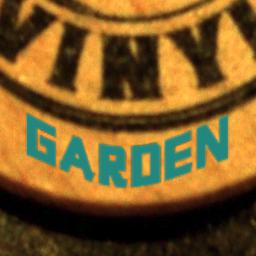}}
& \raisebox{-0.5\height}{\includegraphics[width=1.25\linewidth,height=1.25\linewidth]{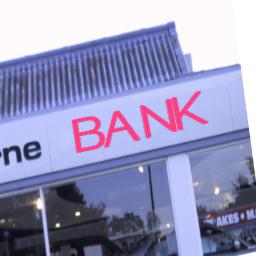}}
& \raisebox{-0.5\height}{\includegraphics[width=1.25\linewidth,height=1.25\linewidth]{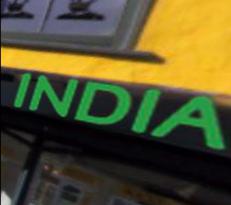}}
\vspace{-2.5 pt}
\\

\raisebox{-0.5\height}{\footnotesize{SF-GAN}} 
& \raisebox{-0.5\height}{\includegraphics[width=1.25\linewidth,height=1.25\linewidth]{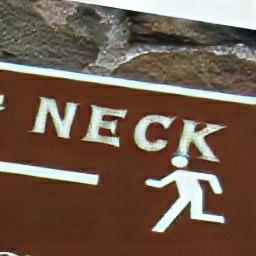}} 
& \raisebox{-0.5\height}{\includegraphics[width=1.25\linewidth,height=1.25\linewidth]{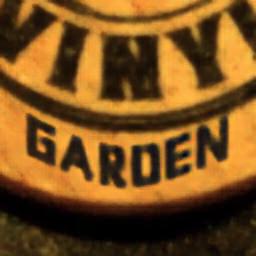}}
& \raisebox{-0.5\height}{\includegraphics[width=1.25\linewidth,height=1.25\linewidth]{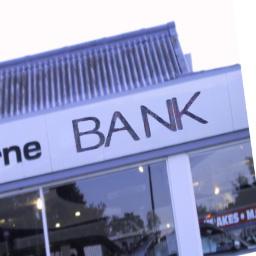}}
& \raisebox{-0.5\height}{\includegraphics[width=1.25\linewidth,height=1.25\linewidth]{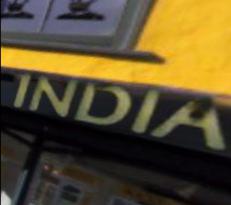}}
\vspace{-2.5 pt}
\\
\vspace{2.5 pt}
\end{tabular}
\caption{Illustration of scene text image synthesis by different GANs: Rows 1-2 are foreground texts and background images as labelled. Rows 3-4 show the images synthesized by ST-GAN and CycleGAN, respectively. Row 5 shows images synthesized by SF-GAN(GS), the output of the geometry synthesizer in SF-GAN (\textit{Composed Image} in Fig. 2). The last row shows images synthesized by the proposed SF-GAN.}
\end{figure}

\noindent
\textbf{Results Analysis:} We use 1 million SF-GAN synthesized scene text images to train scene text recognition models and use the model recognition performance to evaluate the usefulness of the synthesized images. In addition, the SF-GAN is benchmarked with a number of state-of-the-art synthesis techniques by randomly selecting 1 million synthesized scene text images from \cite{jaderberg2014} and randomly cropping 1 million scene text images from \cite{gupta2016} and \cite{zhan2018ver}. Beyond that, we also synthesize 1 million scene text images with random text appearance by using ST-GAN \cite{lin2018}. There are many scene text recognition models \cite{su2014,su2017,tian2016,bshi2018aster,cheng2018aon}, we design an attentional scene text recognizer with a 50-layer ResNet as the backbone network.

For ablation analysis, we evaluate SF-GAN(GS) which denotes the output of the geometry synthesizer (\textit{Composed Image} as shown in Fig. 2) and SF-GAN(AS) which denotes the output of the appearance synthesizer with random geometric alignments.
A baseline SF-GAN (BS) is also trained where texts are placed with random alignment and appearance. The three SF-GANs also synthesize 1 million images each for scene text recognition tests. The recognition tests are performed over four regular scene text datasets ICDAR2013 \cite{icdar2013}, ICDAR2015 \cite{icdar2015}, SVT \cite{wang2011}, IIIT5K \cite{iiit5k} and two irregular datasets SVTP \cite{phan2013} and CUTE \cite{risnumawan2014} as described in \textbf{Datasets}. Besides the scene text recognition, we also perform user studies with Amazon Mechanical Turk (AMT) where users are recruited to tell whether SF-GAN synthesized images are real or synthesized. 

Tables 2 and 3 show scene text recognition and AMT user study results. As Table 2 shows, SF-GAN achieves the highest recognition accuracy for most of the 6 datasets and an up to 3\% improvement in average recognition accuracy (across the 6 datasets), demonstrating the superior usefulness of its synthesized images while used for training scene text recognition models. The ablation study shows that the proposed geometry synthesizer and appearance synthesizer both help to synthesize more realistic and useful image in recognition model training. In addition, they are complementary and their combination achieves a 6\% improvement in average recognition accuracy beyond the baseline SF-GAN(BS). The AMT results in the second column of Table 3 also show that the SF-GAN synthesized scene text images are much more realistic than state-of-the-art synthesis techniques. Note the synthesized images by \cite{jaderberg2014} are gray-scale and not included in the AMT user study.

Fig. 4 shows a few synthesis images by using the proposed SF-GAN and a few state-of-the-art GANs. As Fig. 4 shows, ST-GAN can achieve geometric alignment but the appearance is clearly unrealistic within the synthesized images. The CycleGAN can adapt the appearance of the foreground texts to certain degrees but it ignores real geometry. This leads to not only unrealistic geometry but also degraded appearance as the discriminator can easily distinguish generated images and real images according to the geometry difference.
The SF-GAN (GS) gives the output of the geometry synthesizer, i.e. the \textit{Composed Image} as shown in Fig. 2, which produces better alignment due to good references from the appearance synthesizer. In addition, it can synthesize curve texts due to the use of a thin plate spline transformation \cite{tps}. The fully implemented SF-GAN can further learn text appearance from real images and synthesize highly realistic scene text images. Besides, we can see that the proposed SF-GAN can learn from neighboring texts within the background images and adapt the appearance of the foreground texts accordingly.
\begin{table}
\caption{AMT user study to evaluate the realism of synthesized images. Percentages represent the how often the images in each category were classified as “real” by Turkers.}
\hspace{0.5pt}
\centering 
\begin{tabular}{|l|p{1.4cm}<{\centering}|p{1.4cm}<{\centering}|p{1.4cm}<{\centering}|} 
\hline
Methods & Text & Glass & Hat \\
\hline
Gupta \cite{gupta2016} & 38.0 & - &- \\
Zhan \cite{zhan2018ver} & 41.5 & - & - \\
ST-GAN \cite{lin2018} & 31.6 & 41.7 & 42.6 \\
Real & 74.1 & 78.6 & 78.2 \\
\hline
SF-GAN & 57.7 & 62.0 & 67.3 \\\hline
\end{tabular}
\end{table}

\subsection{Portrait Wearing}
\noindent
\textbf{Data preparation:} We use the dataset CelebA \cite{liu2015} and follow the provided training/test split for portrait wearing experiment. The training set is divided into two groups by using the annotation `glass' and `hat', respectively. For the glass case, one group of people with glasses serve as the real data for matching against in our adversarial settings and the other group without glasses serves as the background. For the foreground glasses, we crop 15 pairs of front-parallel glasses and reuse them to randomly compose with the background images. 
According to our experiment, 15 pairs of glasses as the foreground objects are sufficient to train a robust model. The hat case has the similar setting, except that we use 30 cropped hats as the foreground objects.
\begin{figure}[t]
\begin{tabular}{cccc}
Objects & Faces & ST-GAN & SF-GAN \\\hline
 \vspace{-10 pt}
 & & \\
\vspace{5 pt}
\raisebox{-0.5\height}{\includegraphics[width=0.2\linewidth,height=1.5cm]{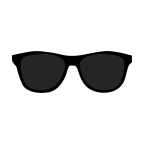}} 
& \raisebox{-0.5\height}{\includegraphics[width=0.2\linewidth,height=1.5cm]{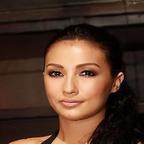}} 
& \raisebox{-0.5\height}{\includegraphics[width=0.2\linewidth,height=1.5cm]{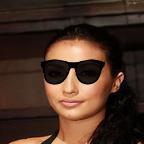}}
& \raisebox{-0.5\height}{\includegraphics[width=0.2\linewidth,height=1.5cm]{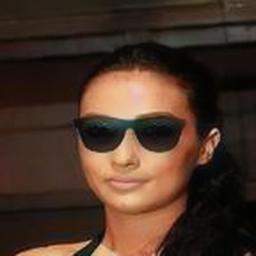}}
\\
\vspace{5 pt}
\raisebox{-0.5\height}{\includegraphics[width=0.2\linewidth,height=1.5cm]{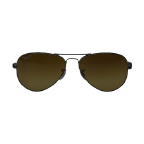}} 
& \raisebox{-0.5\height}{\includegraphics[width=0.2\linewidth,height=1.5cm]{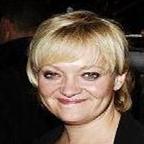}} 
& \raisebox{-0.5\height}{\includegraphics[width=0.2\linewidth,height=1.5cm]{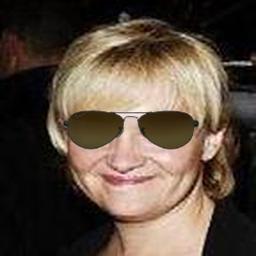}}
& \raisebox{-0.5\height}{\includegraphics[width=0.2\linewidth,height=1.5cm]{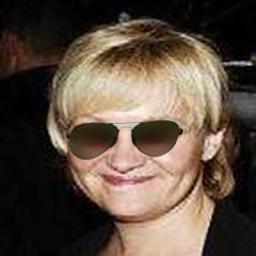}}
\\
\vspace{5 pt}
\raisebox{-0.5\height}{\includegraphics[width=0.2\linewidth,height=1.5cm]{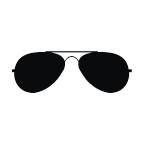}} 
& \raisebox{-0.5\height}{\includegraphics[width=0.2\linewidth,height=1.5cm]{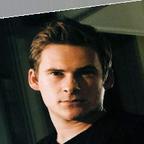}} 
& \raisebox{-0.5\height}{\includegraphics[width=0.2\linewidth,height=1.5cm]{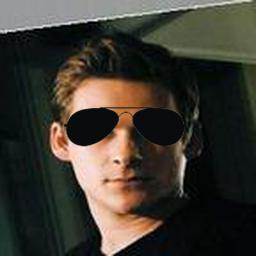}}
& \raisebox{-0.5\height}{\includegraphics[width=0.2\linewidth,height=1.5cm]{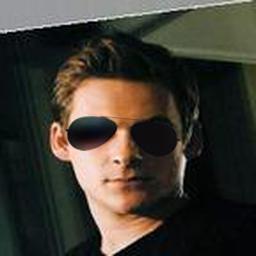}}
\\
\vspace{5 pt}
\raisebox{-0.5\height}{\includegraphics[width=0.2\linewidth,height=1.5cm]{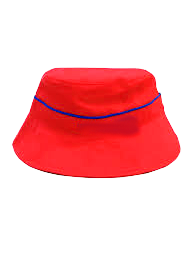}} 
& \raisebox{-0.5\height}{\includegraphics[width=0.2\linewidth,height=1.5cm]{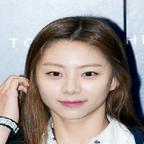}} 
& \raisebox{-0.5\height}{\includegraphics[width=0.2\linewidth,height=1.5cm]{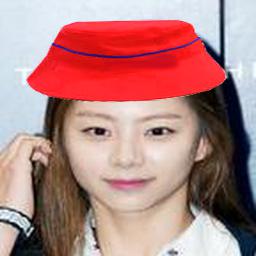}}
& \raisebox{-0.5\height}{\includegraphics[width=0.2\linewidth,height=1.5cm]{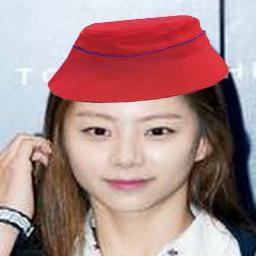}}
\\
\vspace{5 pt}
\raisebox{-0.5\height}{\includegraphics[width=0.2\linewidth,height=1.5cm]{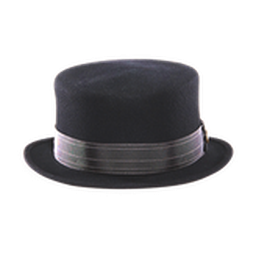}} 
& \raisebox{-0.5\height}{\includegraphics[width=0.2\linewidth,height=1.5cm]{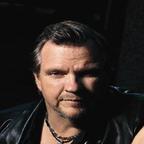}} 
& \raisebox{-0.5\height}{\includegraphics[width=0.2\linewidth,height=1.5cm]{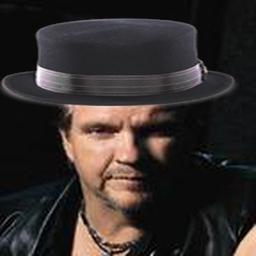}}
& \raisebox{-0.5\height}{\includegraphics[width=0.2\linewidth,height=1.5cm]{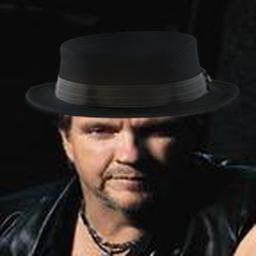}}
\\
\vspace{5 pt}
\raisebox{-0.5\height}{\includegraphics[width=0.2\linewidth,height=1.5cm]{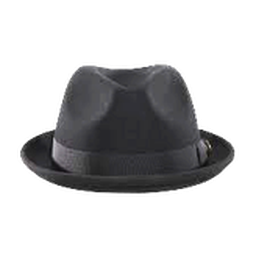}} 
& \raisebox{-0.5\height}{\includegraphics[width=0.2\linewidth,height=1.5cm]{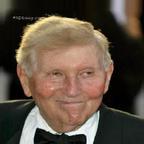}} 
& \raisebox{-0.5\height}{\includegraphics[width=0.2\linewidth,height=1.5cm]{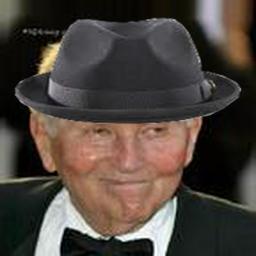}}
& \raisebox{-0.5\height}{\includegraphics[width=0.2\linewidth,height=1.5cm]{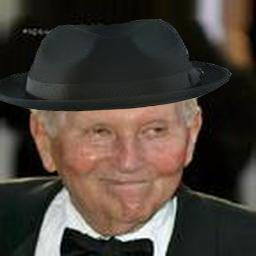}}
\\
\end{tabular}
\caption{Illustration of portrait-wearing by different GANs: Columns 1-2 show foreground hats and glasses and background face images, respectively. Columns 3-4 show images synthesized by by ST-GAN \cite{lin2018} and our proposed SF-GAN, respectively.}
\end{figure}

\noindent
\textbf{Results Analysis:} Fig 5. shows a few SF-GAN synthesized images and comparisons with ST-GAN synthesized images. As Fig. 5 shows, ST-GAN achieves realism in the geometry space by aligning the glasses and hats with the background face images. On the other hand, the synthesized images are unrealistic in the appearance space with clear artifacts in color, contrast and brightness. As a comparison, the SF-GAN synthesized images are much more realistic in both geometry and appearance spaces. In particular, the foreground glasses and hats within the SF-GAN synthesized images have harmonious brightness, contrast, and blending with the background face images. Additionally, the proposed SF-GAN also achieve better geometric alignment as compared with ST-GAN which focuses on geometric alignment only. We conjecture that the better geometric alignment is largely due to the reference from the appearance synthesizer. The AMT results as shown in the last two columns of Table 3 also show the superior synthesis performance of our proposed SF-GAN.

\section{Conclusions}
This paper presents a SF-GAN, an end-to-end trainable network that synthesize realistic images given foreground objects and background images. The SF-GAN is capable of achieving synthesis realism in both geometry and appearance spaces concurrently. The first scene text image synthesis study shows that the proposed SF-GAN is capable of synthesizing useful images to train better recognition models. The second portrait-wearing study shows the SF-GAN is widely applicable and can be easily extend to other tasks. 
We will continue to study SF-GAN for full-image synthesis for training better detection models.

{\small
\bibliographystyle{ieee_fullname}
\bibliography{egbib}
}

\end{document}